\documentclass[10pt,twocolumn,letterpaper]{article}

\usepackage[pagenumbers]{iccv} % To force page numbers, e.g. for an arXiv version

\newcommand{\D}{\mathcal{D}}
\usepackage{xcolor}
\usepackage{colortbl}
\definecolor{lightgray}{gray}{0.85}

\definecolor{newblue}{rgb}{0.21,0.49,0.74}
\usepackage[pagebackref,breaklinks,colorlinks,allcolors=newblue]{hyperref}

\title{CarGait: Cross-Attention based Re-ranking for Gait recognition}

\author{Gavriel Habib
\and
Noa Barzilay
\and
Or Shimshi
\and 
Rami Ben-Ari
\and
Nir Darshan
\and
\mbox{}
\centerline{OriginAI, Israel}
\\
{\tt\small\{gavrielh, noab, ors, ramib, nir\}@originai.co}}

\begin{document}
\maketitle
\begin{abstract}
Gait recognition is a computer vision task that identifies individuals based on their walking patterns. Gait recognition performance is commonly evaluated by ranking a gallery of candidates and measuring the accuracy at the top Rank-$K$. Existing models are typically single-staged, i.e. searching for the probe's nearest neighbors in a gallery using a single global feature representation. Although these models typically excel at retrieving the correct identity within the top-$K$ predictions, they struggle when hard negatives appear in the top short-list, leading to relatively low performance at the highest ranks (\eg, Rank-1). In this paper, we introduce CarGait, a Cross-Attention Re-ranking method for gait recognition, that involves re-ordering the top-$K$ list leveraging the fine-grained correlations between pairs of gait sequences through cross-attention between gait strips. This re-ranking scheme can be adapted to existing single-stage models to enhance their final results. We demonstrate the capabilities of CarGait by extensive experiments on three common gait datasets, Gait3D, GREW, and OU-MVLP, and seven different gait models, showing consistent improvements in Rank-1,5 accuracy, superior results over existing re-ranking methods, and strong baselines.
\end{abstract}    
\section{Introduction}
\label{sec:intro}

Gait recognition (GR) identifies individuals based on their walking patterns, utilizing features like body shape, stride length, gait cycle dynamics, and limb movements. GR is applicable in various fields, including healthcare~\cite{majumder2018simple}, criminal investigations~\cite{bouchrika2011using}, surveillance~\cite{parashar2023real}, and sport~\cite{xu2024new}. The complexity in GR arises from several factors, \eg diverse camera views, occlusions, changing in clothing, or carrying of bags, that alter the shape and the observed gait dynamics. 

Gait recognition is typically framed as a retrieval task, where there is a gallery of encoded gait sequences from various individuals, each associated with an identity. Given a new probe sequence, the objective is to retrieve the sequences from the gallery that match the probe, based on their gait embeddings. 

The performance of gait recognition models is evaluated using Rank-$K$ accuracy, where the gallery is ranked in ascending distance order (descending similarity) from the probe. Ideally, the probe's matching identity should be at the top of the list (high Rank-1 accuracy). In real-world applications like security and surveillance, accurately identifying an individual at Rank-1 is important for fast and precise decision-making, minimizing the need for additional verification steps. It further facilitates user experience and better demonstrates the model's robustness to varying conditions, like changes in lighting, clothing, or walking styles.

\begin{figure}[!t]
\centering
\includegraphics[width=0.65\linewidth]{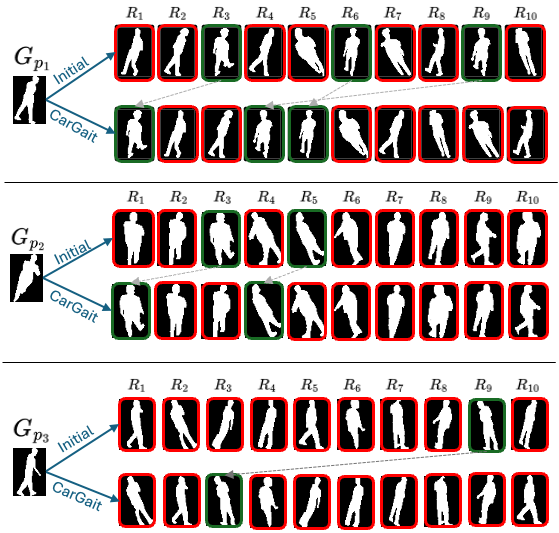}
\caption{Examples of top-10 retrievals (from left to right) of probe sequences from the {\it Gait3D}~\cite{Zheng_2022_CVPR} and {\it OU-MVLP}~\cite{takemura2018multi} datasets, before and after applying our CarGait re-ranking method. To simplify, each gait sequence is shown by a single image. Top row: without re-ranking (Initial), Bottom row: with CarGait re-ranker. Green rectangles indicate correct identity (true-positives), while the red are incorrect recognition (false-positives). CarGait improves Rank-1 and Rank-5 by initial list re-ordering.}
\label{fig:visual_results}
\end{figure}

Current gait recognition models~\cite{chao2019gaitset, fan2020gaitpart, fan2023opengait, fan2024skeletongait} operate in a single stage, encoding gait sequences into a single and often referred to as a {\it global} feature, which allows for efficient search in scale and ranking within a large dataset.
In recent years, significant efforts have been made to enhance global gait representation~\cite{fan2023exploring, fan2023learning, fan2024skeletongait, habib2024watch, ye2024biggait}, by changing architectures to transformers~\cite{fan2023exploring} or combining different modalities~\cite{fan2024skeletongait}, yet overlooking a two-stage approach commonly employed in various domains, \eg image retrieval~\cite{shao2023global}, visual place recognition~\cite{zhu2023r2former, zhang2023etr, tzachor2024effovpr}, and person re-identification~\cite{zhong2017re, bai2019re, zhou2022moving}. Two-stage methods incorporate a second stage, following the {\it initial} stage, referred as {\it re-ranking}.
Re-ranking is the process of re-ordering the top $K$ predictions, which are initially retrieved in a global stage from a large dataset and ranked based on their similarity to the probe. A re-ranker often leverages additional information~\cite{zhong2017re, chen2016uncooperative} and can afford higher computational costs, since it operates on a relatively short list. 

While recent GR models generally perform well across various datasets~\cite{Zheng_2022_CVPR, Zhu_2021_ICCV, takemura2018multi}, they often face challenges in achieving high Rank-1 accuracy. This can be attributed to distortions in the model's typical inputs, such as silhouettes or skeletons, as well as the presence of hard-negatives in the gallery (identities with gait patterns similar to the probe). Moreover, the model's reliance on a single global representation limits its discrimination capability.
There is often a large gap in retrieving the correct identity at the first rank than within the top-5. For example, GaitPart model~\cite{fan2020gaitpart} Rank-1 accuracy on the Gait3D dataset~\cite{Zheng_2022_CVPR} is 28.2\%, while its Rank-5 is 47.6\% (see more examples in~\cref{tab:results}, under "initial" column). This underscores the potential to enhance Rank-1 accuracy by effectively re-ordering the top results, even within the top-5 shortlist. This paper addresses this re-ranking problem. 

Gait strips are spatio-temporal aggregated units (weakly) associated with spatial parts of the human body~\cite{wang2022gaitstrip, fan2020gaitpart}. The distance between two gait features is often computed as an average over corresponding strip distances~\cite{fan2023exploring, chao2019gaitset, fan2020gaitpart}. This implies that strips carry gait information in a fine-grained manner. Since the top-$K$ results at the global stage is populated with hard-negative cases, it is natural to assume that a fine-grained comparison can improve over the initial global feature based ranking.

In this paper, we present a re-ranking method for gait recognition that can be integrated with existing single-stage gait models. Building on a pre-trained gait recognition model, we propose CarGait, a Cross-Attention based Re-ranking approach designed to enhance identity recognition by capturing fine-grained correlations between gait sequences. This is achieved through cross-attention between different gait strips of two sequences, the probe and each of its top-$K$ ranked candidates. Leveraging a metric learning approach, we map the original embedding space into a new one (of the same dimension), where probe-candidate distances are adjusted to improve Rank-1 and Rank-5 accuracy (see~\cref{fig:visual_results}). This enhancement stems from learning fine-grained and meaningful interactions between body strips, allowing for better differentiation between subtle variations of the same identity (positives) and hard negatives appearing at the top of the ranked list.

We demonstrate the capabilities of CarGait through extensive evaluations on three common gait datasets, Gait3D, GREW, and OU-MVLP, and seven gait models, with additional baseline, showing consistent improvements in Rank-1 and 5 accuracy. ~\Cref{fig:spider_chart} depicts the Rank-1 improvements by CarGait.

In summary, our key contributions are as follows:
\begin{itemize}
\item Targeting the underexplored task of re-ranking in gait recognition, we introduce a tailored metric learning to learn new discriminative embeddings.
\item Proposing a fine-grained refinement of similarities via pairwise probe-candidate interactions, we learn conditioned representations across all gait strips of two individuals using cross-attention.
\item CarGait consistently outperforms baselines and state-of-the-art re-ranking methods in person re-identification across diverse models and challenging datasets.
\end{itemize}
\section{Related Work}
\label{sec:related_work}

Gait recognition models are commonly classified into two categories: model-based and appearance-based approaches. Model-based approaches~\cite{liao2020model, teepe2021gaitgraph, teepe2022towards, zhang2023spatial, guo2023physics, fu2023gpgait} try to recognize walking patterns using the estimated structure of the human body, such as 2D or 3D pose, while appearance-based approaches~\cite{chao2019gaitset, fan2020gaitpart, fan2023opengait} extract gait features directly from RGB images or binary silhouette sequences. Although model-based approaches are theoretically robust to changes in clothing and carrying objects, they tend to under-perform appearance-based approaches on in-the-wild benchmarks, \eg Gait3D~\cite{Zheng_2022_CVPR} and GREW~\cite{Zhu_2021_ICCV}. This disparity is likely due to the challenges in estimating body parameters in low-resolution videos. In this work, we focus on appearance-based approaches due to their demonstrated superiority.

\subsection{CNNs and Transformers}
Although transformers have demonstrated superior performance in numerous computer vision tasks, the majority of common gait recognition models are still based on CNNs ~\cite{chao2019gaitset,fan2020gaitpart,lin2021gait,fan2023opengait, fan2024skeletongait}. However, some recent methods leverage the capabilities of Transformers for gait recognition~\cite{mogan2022gait,fan2023exploring,ma2024learning}. Our method is compatible with both architectures and can be learned on top of each. We specifically demonstrate this capability on the following CNN~\cite{chao2019gaitset, lin2021gait, fan2020gaitpart, fan2023opengait, fan2024skeletongait} and Transformer architectures~\cite{fan2023exploring}. 
The self-attention mechanism~\cite{vaswani2017attention, dosovitskiy2020image} is widely used in gait recognition~\cite{mogan2022gait,fan2023exploring,ma2024learning} to emphasize key spatial areas or time slots within a single sequence. In contrast, we have created a cross-attention component specifically designed to learn the relationships between two distinct gait sequences.

%-------------------------------------------------------------------------
\subsection{Cross-Attention}
Cross attention has been used in various applications~\cite{ahn2023star, levy2024data, liu2024video}. Specifically, in gait recognition, Cui \etal~\cite{cui2023multi} used cross-attention to combine features of a gait sequence from different modalities, such as silhouettes and skeletons for global single-stage ranking. In our study, we introduce a cross-attention method specifically designed for re-ranking. 

While previous models utilized the corresponding gait strips to optimize the feature space to rank the entire gallery directly~\cite{chao2019gaitset, lin2021gait, fan2020gaitpart, fan2023opengait, fan2023exploring, fan2024skeletongait}, CarGait takes a different approach. It maps the global feature maps from a pre-trained single-stage model into a new feature space, by applying cross-attention across \textit{all strips} of the probe and its top candidates. This transformation produces refined feature representations that enhance the re-ranking process.

%-------------------------------------------------------------------------
\subsection{Re-ranking}
\label{sec:re-ranking}
Re-ranking is the process of refining or re-ordering an initial list of results to enhance the accuracy and relevance of the final ranked list. It is widely used in various applications, such as text retrieval~\cite{nogueira2019passage,sun2023chatgpt} and different applications of image retrieval \cite{tan2021instance, shao2023global,zhu2023r2former}. However, these methods are often not designed to address spatio-temporal "instance" matching, as in gait recognition. In this context, Gordo \etal ~\cite{gordo2020attention} suggested the "query expansion method", which was originally designed for one-stage image retrieval and was later used by~\cite{qu2023learnable} for re-ranking. Our method is different as it conducts a pairwise interaction and impacts not only the probe representation but the candidate as well. 

A closer domain to gait recognition in re-ranking is person re-identification (reID), which focuses on matching images of the same individual across different viewpoints using visual cues like clothing. Notable approaches include $k$-reciprocal encoding (KR)~\cite{zhong2017re}, ECN~\cite{sarfraz2018pose}, and a specialized feature-learning method~\cite{zhang2023specialized} that leverages the relative structure of the top-ranked list in the feature space. In this context, Bai \etal~\cite{bai2019re} introduced a re-ranking approach for object retrieval and reID by incorporating metric fusion and capturing the geometric structure of multiple data manifolds. LBR~\cite{luo2019spectral} formulates data as a graph to optimize group-wise similarities, while GCR~\cite{zhang2023graph} employs a graph-based strategy that refines the global feature representations by aligning them across similar samples. Instead, CarGait introduces a {\it fine-grained} and \textit{pairwise} comparison between the probe and each individual candidate. This distinct approach is better suited for gait recognition and performs better than common re-rankers, particularly in scenarios where positives are rare in the gallery~\cite{Zhu_2021_ICCV, Zheng_2022_CVPR}, even when there is only a single positive sample (as demonstrated in~\cref{tab:reranking_comparison_oumvlp}).

Unlike re-ranking methods that often focus on refining the similarity matrix of the top-$K$ samples~\cite{zhong2017re, sarfraz2018pose, luo2019spectral}, CarGait modifies the gait representations at the re-ranking stage, using detailed comparisons through cross-attention. 

Despite their potential to boost accuracy, re-ranking methods are rarely used in gait recognition.
In this regard, Chen \etal~\cite{chen2016uncooperative} proposed a re-ranking method based on \textit{engineered features} such as Gait Energy Image (GEI) or Active Energy Image (AEI). They enhance robustness against external factors like clothing changes, view angle, and walking speed by transferring the GEI into a new feature space through sparse coding.  In contrast, CarGait leverages a \textit{deep learning} architecture that integrates feature learning into the re-ranking process. Our approach employs \textit{cross-attention} to capture fine-grained interactions between pairs of probe and candidate feature maps, refining the ranking process.

In summary, most existing re-ranking methods are designed for image retrieval or person re-identification, focusing on refining rankings based on the spatial or relative arrangement of samples in the initial embedding space. In contrast, CarGait generates a new embedding space specifically tailored for gait spatio-temporal re-ranking. To highlight the advantages of CarGait, we conduct extensive comparisons with various approaches from these domains, KR~\cite{zhong2017re}, LBR~\cite{luo2019spectral}, and GCR~\cite{zhang2023graph}, demonstrating its effectiveness in re-ranking for gait recognition.
%-------------------------------------------------------------------------

\section{Method}
\label{sec:method}

\begin{figure}[!t]
\centering
\includegraphics[width=\linewidth]{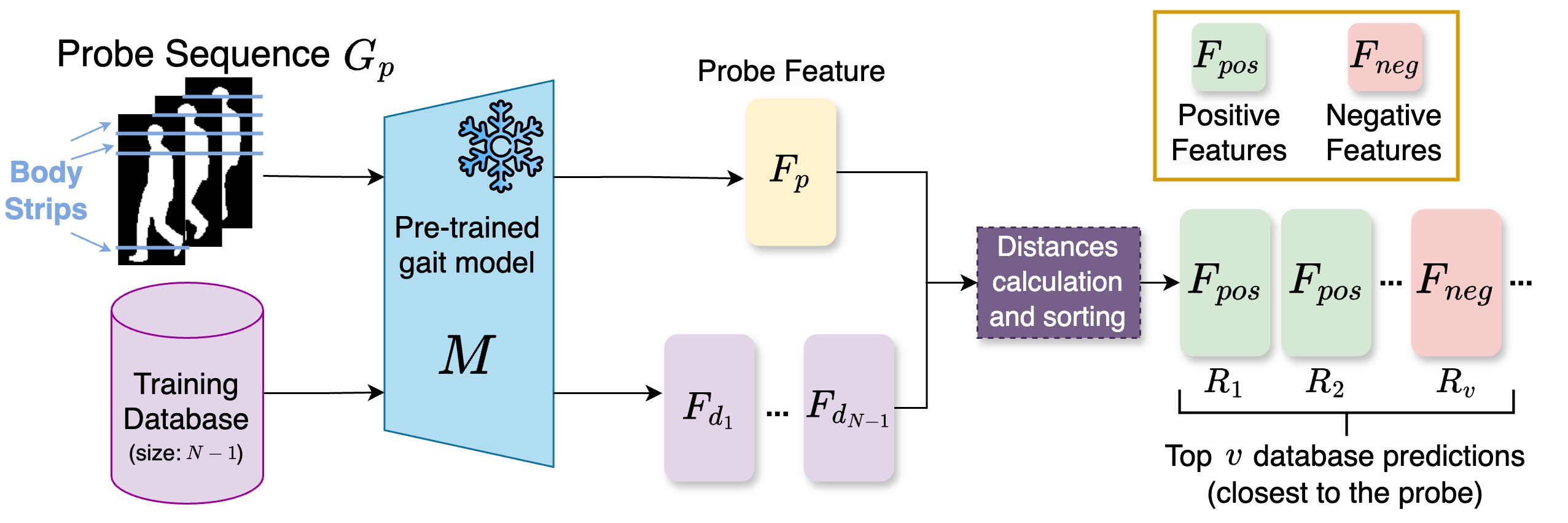}
\caption{{\bf Train set generation:} a pre-trained gait model $M$ is used as a feature extractor. A training dataset of gait probe feature maps $F_{p}$, with their nearest $v$ candidates, is constructed. Within this top-$v$ list, some features may be positives (\ie, sharing the same identity as the probe, shown in green), while others may be negatives (shown in red).}
\label{fig:train_set_generation}
\end{figure}

\begin{figure*}[!t]
\centering
\includegraphics[width=0.79\linewidth]{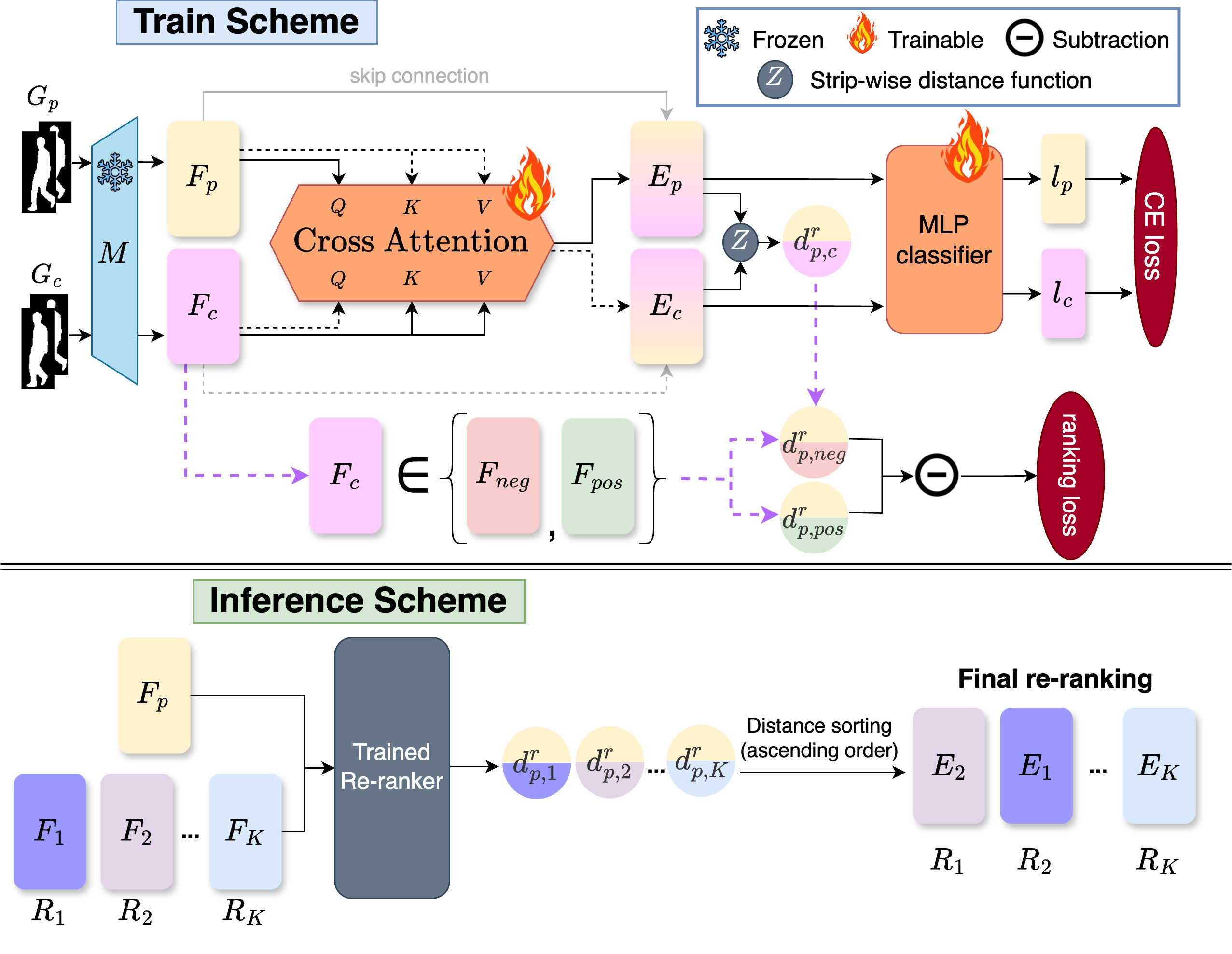}
\caption{An overview of CarGait method. {\bf Train Scheme:} a strip-wise multi-head cross-attention based re-ranker is trained with ranking and cross-entropy losses, learning the part relations between pairs of gait sequences. Practically, the cross-attention module is applied twice for each probe-candidate pair (illustrated by the solid and dashed lines). {\bf Inference Scheme:} re-ranking is achieved by sorting the probe's top $K$ candidate predictions in ascending order by their new distances to the probe, as determined by the trained re-ranker with $d^r_{p,c}$.}
\label{fig:method}
\end{figure*}

In this section, we introduce CarGait, a novel re-ranking method for gait recognition. ~\Cref{fig:method} presents an overview of CarGait with its training phase as well as the inference scheme. Our re-ranking method is based on cross-attention between the probe and each candidate in its top-$K$ global ranking results. Let us start by denoting a pair of gait sequences, $G_{p}$ and $G_{c}$, as the {\it probe} and {\it candidate} sequence respectively, each consisting of a set of silhouette frames and optionally also skeletons. ~\Cref{fig:train_set_generation} illustrates the process of generating the training data for the re-ranker. 

We now define the gait feature maps derived from \( G_{p} \) and \( G_{c} \) after being processed by a pre-trained model \( M \), yielding \( M(G_p, G_c) = \{F_{p}, F_{c}\} \). Here, \( F_{p}, F_{c} \in \mathbb{R}^{s\times d} \) are the extracted feature maps, where \( s \) represents the number of horizontal body \textit{strips}, and \( d \) denotes the feature dimension. These feature maps are obtained after temporal aggregation within the backbone.  

Next, we compute multi-head cross-attention between \( F_{p} \) and \( F_{c} \) (see~\cref{fig:method}, top-view), where \( F_{p} \) serves as the Query and \( F_{c} \) as both the Key and Value (solid lines), producing a new feature map \( E_{p} \in \mathbb{R}^{s \times d} \). Similarly, we obtain \( E_{c} \in \mathbb{R}^{s\times d} \) by performing the reverse cross-attention, using \( F_{c} \) as the Query and \( F_{p} \) as the Key and Value (dashed lines).  
\begin{figure}[!t]
\centering
\includegraphics[width=\linewidth]{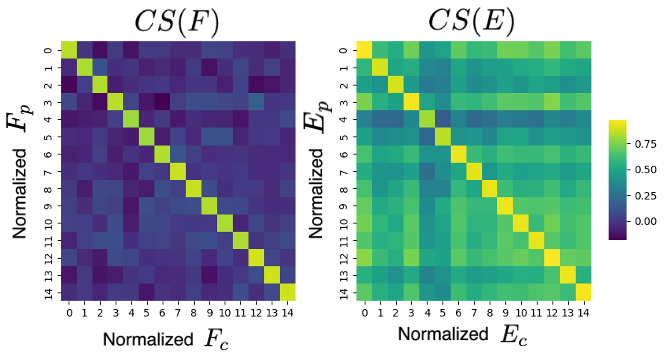}
\caption{Cosine similarities ($CS$) between L2-normalized features of $s$ body strips (here, $s=15$) in two gait sequences of the same identity, in the initial global space $F$ and the re-ranker space $E$. Each row/column represents a strip. In $CS(F)$, the features are distinct, indicating low correlation between different body strips (blue colors off-diagonal). In the re-ranker space $CS(E)$, that leverages cross-attention, higher correlation between strips are observed (green/yellow colors), indicating cross-strip interactions learned through the cross-attention. Samples are from the {\it Gait3D} dataset~\cite{Zheng_2022_CVPR}.}
\label{fig:body_strips}
\end{figure}

To preserve information from the pre-trained model, we incorporate a residual (skip) connection between \( F_{p} \) and \( E_{p} \), as well as between \( F_{c} \) and \( E_{c} \). This ensures that the re-ranker is initialized with the pre-trained feature space while refining the representations. In practice, the cross-attention module takes two \textit{distinct} feature maps as input and generates two \textit{conditioned} representations. Each strip in \( E_{p} \) is now influenced by its attention relationships with all the strips in the candidate.

We now derive a new metric space, with modified probe and candidate representations, for re-ranking. The distance is computed as the average euclidean distance between corresponding strip features {\it after} cross-attention, namely, $d^r_{p,c} = \mathcal{Z}(E_p,E_c)$, where the function $\mathcal{Z}$ from~\cite{fan2023exploring, chao2019gaitset, fan2020gaitpart} maps two given representations to a distance. We train our re-ranker with two loss objectives, ranking and classification (see~\cref{eq:model_losses}). For the classification loss, we add a trainable MLP, applied on top of $E_{p}$ and $E_{c}$ (see~\cref{fig:method} top-view). The classification loss functions as a regularization term, preserving identity information within the learned representations and contributing to improved performance.

\Cref{fig:body_strips} illustrates the impact of our cross-attention module. We compute the cosine similarity matrix between strips of a probe and a positive candidate, \textit{i.e} from the same identity. For comparison, we show the strip correlations from the initial global features $F_{p}, F_{c}$, namely $CS(F)$ vs.  $CS(E)$, based on $E_{p}$ and $E_{c}$ strips after our learnable cross-attention module for re-ranking, indicating CarGait. In both cases, matched body strips have higher similarity (diagonals), while in CarGait new interactions are learned between different body strips with the cross-attention module (brighter colors on off-diagonal blocks).

We train our re-ranker for each global model $M$, keeping $M$'s layers frozen. First, we generate a dataset $\D$ comprising the feature map representations ($F_{p},F_{c}$) of all training samples along with their top-$v$ closest predictions, as illustrated in~\cref{fig:train_set_generation}. Then we train CarGait modules (see~\cref{fig:method}) with an objective to improve the ranked list. 

%%%%%%%%%%%%%%%%%%%%%%%%%%%%%%%
{\bf Loss:} We optimize our re-ranking module using two loss objectives: ranking loss and classification loss. The ranking loss is a metric-based loss that penalizes the model whenever a negative sample is positioned closer to the probe than a positive one, in Euclidean space. The penalty is proportional to the relative distance difference between the probe-negative and probe-positive pairs. The classification loss can be viewed as a regularization term ensuring that identity information is effectively preserved within the learned representations. This helps maintain discriminative features while refining the ranking process.

Considering a triplet of gait feature maps for probe ($p$) and corresponding positive ($pos$ - same identity) and negative ($neg$) samples, denoted by $\{F_{p_i}, F_{pos_i}, F_{neg_i}\}$ sampled from $\D$, the ranking loss is given by:
\begin{equation}
\begin{aligned}
& \mathcal{L}_{i}^*\ = -log[\sigma(d^r_{p_i, neg_i} - d^r_{p_i, pos_i})] \\
& \mathcal{L}_{i} = \begin{cases} 
                   \beta{\mathcal{L}_{i}^*}\, & \text{if } d^r_{p_i, neg_i} \geq d^r_{p_i, pos_i} \\
                   \mathcal{L}_{i}^*\, & \text{otherwise}
                 \end{cases} \\
& \mathcal{L}_{ranking} = \sum_{i}\mathcal{L}_{i}
\end{aligned}
\label{eq:ranking_loss}
\end{equation}
where $\sigma$ is a sigmoid function and $d^r_{\cdot,\cdot}$ indicates distance in the new re-ranking space. For sake of effectiveness, during CarGait training, we downscale the loss of the triplets that are correctly ranked (\ie $d^r_{p_i, neg_i} \geq d^r_{p_i, pos_i}$) by a scale factor $\beta < 1$.

The classification loss ($\mathcal{L}_{CE}$) is a standard multi-class cross-entropy loss~\cite{de2005tutorial}, applied to all training samples. To compute this loss, the attended representation $E_{i}$ is fed into an MLP classifier, producing a logits vector $l_{i}$ with a size matching the number of training classes $C$. Eventually, the losses are linearly combined:
\begin{align}
\mathcal{L} & = \mathcal{L}_{ranking} + \alpha\mathcal{L}_{CE}
\label{eq:model_losses}
\end{align}
where $\alpha$ is a standard weighting hyper-parameter.

{\bf Train Stopping Criteria: }
We use $\D_{val}$, a dataset constructed in the same manner as $\D$ but derived from the validation set, for stopping criteria. Specifically, the ranking loss described in~\cref{eq:ranking_loss} is calculated each $T_{val}$ iterations during the training, and eventually the checkpoint with the minimum loss is chosen.

{\bf Inference:} 
For a given probe sequence $G_{p}$, the top-$K$ nearest sequences are first retrieved using a given pre-trained model. The re-ranker is then applied to all pairs $(F_p,F_c)$ within the top-$K$ list, to compute the updated distances ($d^r_{p,1}$, ..., $d^r_{p,K}$). The re-ranking is then achieved by re-ordering the top-$K$ gallery sequences based on their new distances to the probe sequence, in ascending order.
\section{Evaluation}
\label{sec:evaluation}

\subsection{Gait recognition models}
We evaluate CarGait on six silhouette based gait recognition models with a wide range of performance levels: SwinGait-3D, DeepGaitV2-P3D~\cite{fan2023exploring}, GaitBase~\cite{fan2023opengait}, GaitSet~\cite{chao2019gaitset}, GaitGL~\cite{lin2021gait}, and GaitPart~\cite{fan2020gaitpart}, and on a combined silhouette-skeleton SoTA model of SkeletonGait++~\cite{fan2024skeletongait}. These models are then tested across three different benchmarks\footnote{For comparison, we chose the models that offer publicly available code and checkpoints.}. For each model, we train a re-ranker, adjusting the input size according to the feature map dimensions obtained from the pre-trained model. All other hyper-parameters, such as the number of attention heads, remain unchanged.

{\tt SwinGait-3D}~\cite{fan2023exploring} combines convolutional layers followed by transformer blocks on shifted windows.

{\tt DeepGaitV2-P3D (DGV2-P3D)}~\cite{fan2023exploring} and {\tt GaitBase}~\cite{fan2023opengait} are based on deep ResNet, aiming to generalize better for in-the-wild scenarios.

{\tt GaitSet}~\cite{chao2019gaitset} is trained on a set of silhouettes, {\tt GaitPart}~\cite{fan2020gaitpart} emphasizes the importance of body parts, and {\tt GaitGL}~\cite{lin2021gait} fuses global and local information. All three are CNN-based.

{\tt SkeletonGait++ (SG++)}~\cite{fan2024skeletongait} combines silhouette and skeleton features in a multi-branch architecture.

%%%%%%%
\subsection{Datasets}
\label{sec:datasets}
We evaluate CarGait on three well-known gait datasets, {\it Gait3D}~\cite{Zheng_2022_CVPR}, {\it GREW} \cite{Zhu_2021_ICCV}, and {\it OU-MVLP}~\cite{takemura2018multi}. Our experiments strictly adhere to the official evaluation protocols.

{\it Gait3D}~\cite{Zheng_2022_CVPR} contains 25,309 sequences of 4,000 identities recorded from 39 cameras in a supermarket. The training set includes 3,000 subjects, while the test set includes the remaining 1,000 subjects.

{\it GREW}~\cite{Zhu_2021_ICCV} is a large-scale dataset containing 128,671 sequences of 26,345 subjects captured from 882 cameras in the wild. The training set contains 20,000 subjects, while the test set comprises 6,000 subjects.

{\it OU-MVLP}~\cite{takemura2018multi} is a large indoor dataset captured in a controlled environment from 14 viewing angles. It includes 10,307 subjects, with 5,153 used for training and the remaining 5,154 used for testing.

We use the {\tt typewriter} font for model names and {\it italic} font for dataset names for better distinction.

\subsection{Implementation details}
\label{sec:implementation_details}
The implementations of the gait models and checkpoints for academic datasets are publicly available in the OpenGait codebase~\cite{Fan_2023_CVPR}. In each dataset, the input is a silhouette sequence where the silhouettes are resized to 64$\times$44. In SkeletonGait++ model~\cite{fan2024skeletongait}, additional skeleton-based information of size 2$\times$64$\times$44 is added. The training set is divided into separate training and validation sets by designating the last 10\% training identities as the validation set.

We adopt the AdamW optimizer~\cite{loshchilov2017decoupled} with a learning rate of 1e-5 and a weight decay of 1e-2. In each training iteration, a batch size of 32$\times$4 was used\footnote{For DeepGaitV2-P3D the batch size was reduced to 16$\times$4, and for SwinGait-3D and SkeletonGait++ to 8$\times$4, due to memory issues.}. The re-ranker was trained using the loss function described in~\cref{eq:model_losses} with $\alpha = 0.01$, and with $\beta = 0.1$ in the ranking loss (see~\cref{eq:ranking_loss}). The multi-head cross-attention module was designed as a single block with $8$ attention heads and with a hidden dimension of $256D$. For creating the dataset $\D$, the top $v=30$ predictions were considered. The validation loss was calculated every $T_{val}$ = 10,000 training iterations, and the total number of iterations was set to 100,000. During inference, the re-ranking process was applied to the top $K=10$ predictions of each probe sequence. Our experiments were conducted on four NVIDIA A100 GPUs for training, each with 40 GB of memory. The average training time per experiment was 16 hours, with inference time of approximately 6.5 milliseconds per probe on a single GPU. Further runtime analysis is shown in the appendix.

\subsection{Results}
\label{sec:results}

\begin{figure}[!t]
\centering
\includegraphics[width=0.95\linewidth]{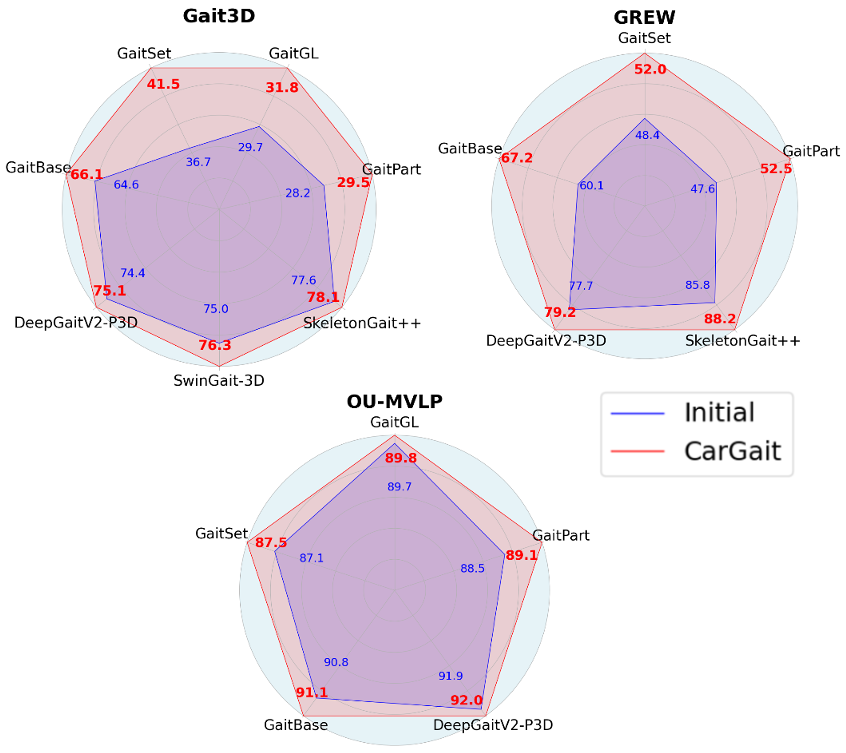}
\caption{Comparison of Rank-1 accuracy of different models before (in blue) and after CarGait re-ranker (in red), across three different benchmarks.}
\label{fig:spider_chart}
\end{figure}

We show the re-ranker results applied on top of various gait recognition models for multiple datasets in~\cref{tab:results}. We compare the results obtained by CarGait applied on the top-10 list, with those of the original pre-trained models before re-ranking (referred as {\it initial} results). A spider-plot visualization of Rank-1 results is shown in~\cref{fig:spider_chart}, along with an illustration of the re-ranking depicted in~\cref{fig:visual_results}. CarGait demonstrates consistent enhancements in Rank-1 accuracy across various models and datasets, highlighting the effectiveness of our approach in improving global single-stage models in gait recognition. For example, in {\tt GaitBase} on {\it GREW} dataset, the initial Rank-1 of 60.1\% is improved to 67.2\%. Note that since our re-ranker is designed to re-arrange a given list of top-10 results, it also improves Rank-5 accuracy. For instance, in {\tt SwinGait-3D} on {\it Gait3D} dataset, the initial Rank-5 of 86.7\% is improved to 88.6\%.

Notably, CarGait shows greater improvements when applied to {\it Gait3D} and {\it GREW} datasets, compared to {\it OU-MVLP}. We suggest that these differences are due to the initial results achieved on each dataset. While {\it Gait3D} and {\it GREW} are challenging datasets captured in-the-wild, {\it OU-MVLP} is an indoor dataset collected in a controlled environment. As a result, the Rank-1 performances on {\it OU-MVLP} appears to be saturated even before applying CarGait, leaving less potential for improvement.

Next, we compare CarGait to three existing re-rankers~\cite{zhong2017re, luo2019spectral, zhang2023graph} across different benchmarks, as presented in~\cref{tab:reranking_comparison_g3d},~\ref{tab:reranking_comparison_oumvlp} (and in Tab.\ $1$ in the appendix). CarGait consistently outperforms other re-rankers~\cite{zhong2017re, luo2019spectral, zhang2023graph} across all benchmarks. We attribute this to its modeling, addressing gait recognition and by capturing probe-candidate internal relationships through strip-wise cross-attention, rather than merely re-ordering the top-$K$ gallery predictions based on global feature similarities. In some scenarios, particularly when positives are rare in the gallery, other methods may even degrade the initial single-stage results, whereas CarGait continues to enhance them (see~\cref{tab:reranking_comparison_oumvlp}).

\Cref{tab:inference_time_analysis} presents the runtime analysis. While CarGait adds some complexity and requires training, it is significantly faster than other re-ranking methods at inference.

\begin{table*}[ht]
\centering
\small
\begin{tabular}{@{}c|c|cccccc|cccc|cc}
\toprule
 & & \multicolumn{6}{c|}{\it Gait3D} & \multicolumn{4}{c|}{\it GREW} & \multicolumn{2}{c}{\it OU-MVLP} \\
  & & \multicolumn{2}{c}{R1} & \multicolumn{2}{c}{R5} & \multicolumn{2}{c|}{mAP} & \multicolumn{2}{c}{R1} & \multicolumn{2}{c|}{R5} & \multicolumn{2}{c}{R1} \\
Method & Publication & Initial & CG & Initial & CG & Initial & CG & Initial & CG & Initial & CG & Initial & CG \\ \midrule
\tt{GLN}~\cite{hou2020gait} {}& ECCV 20 & 31.4 & * & 52.9 & * & 24.74 & * & - & - & - & - & 89.2 & * \\ 
\tt{CSTL}~\cite{huang2021context} {}& ICCV 21 & 11.7 & * & 19.2 & * & 5.59 & * & - & - & - & - & 90.2 & * \\ 
\tt{GaitGCI}~\cite{dou2023gaitgci} {}& CVPR 23 & 50.3 & * & 68.5 & * & 39.50 & * & 68.5 & * & 80.8 & * & 92.1 & * \\ 
\tt{DANet}~\cite{ma2023dynamic} {}& CVPR 23 & 48.0 & * & 69.7 & * & - & * & - & - & - & - & 90.7 & * \\ 
\tt{HSTL}~\cite{wang2023hierarchical} {}& ICCV 23 & 61.3 & * & 76.3 & * & 55.48 & * & 62.7 & * & 76.6 & * & 92.4 & - \\ 
\tt{DyGait}~\cite{wang2023dygait} {}& ICCV 23 & 66.3 & * & 80.8 & * & 56.40 & * & 71.4 & * & 83.2 & * & - & - \\
\tt{VPNet}~\cite{ma2024learning} {}& CVPR 24 & 75.4 & * & 87.1 & * & - & * & 80.0 & * & 89.4 & * & 92.4 & * \\ \midrule
\tt{GaitPart}~\cite{fan2020gaitpart} & CVPR 20 & 28.2 & \textbf{29.5} & 47.6 & \textbf{48.5} & 21.58 & \textbf{22.73} & 47.6 & \textbf{52.5} & 60.7 & \textbf{67.5} & 88.5 & \textbf{89.1} \\
\tt{GaitGL}~\cite{lin2021gait} & ICCV 21 & 29.7 & \textbf{31.8} & 48.5 & \textbf{51.0} & 22.29 & \textbf{23.55} & 47.3 & * & 63.6 & * & 89.7 & \textbf{89.8} \\
\tt{GaitSet}~\cite{chao2019gaitset} {} & AAAI 19 & 36.7 & \textbf{41.5} & 58.3 & \textbf{62.1} & 30.01 & \textbf{32.97} & 48.4 & \textbf{52.0} & 63.6 & \textbf{68.0} & 87.1 & \textbf{87.5} \\
\tt{GaitBase}~\cite{fan2023opengait} & CVPR 23 & 64.6 & \textbf{66.1} & 81.5 & \textbf{82.8} & 55.29 & \textbf{57.66} & 60.1 & \textbf{67.2} & 75.5 & \textbf{78.5} & 90.8 & \textbf{91.1}  \\
\tt{DGV2-P3D}~\cite{fan2023exploring} {} & ArXiv 23 & 74.4 & \textbf{75.1} & \textbf{88.0} & 87.5 & 65.76 & \textbf{66.89} & 77.7 & \textbf{79.2} & 87.9 & \textbf{88.7} & 91.9 & \textbf{92.0} \\
\tt{SwinGait3D}~\cite{fan2023exploring} {}& ArXiv 23 & 75.0 & \textbf{76.3} & 86.7 & \textbf{88.6} & 66.69 & \textbf{67.59} & 79.3 & * & 88.9 & * & - & - \\
\tt{SG++}~\cite{fan2024skeletongait} {}& AAAI 24 & 77.6 & \textbf{78.1} & 89.4 & \textbf{90.4} & 70.30 & \textbf{70.86} & 85.8 & \textbf{88.2} & 92.6 & \textbf{94.6} & - & - \\
\bottomrule
\end{tabular}

\caption{Improvement in Rank-$K$ accuracy [\%] and mAP with CarGait re-ranker, over different {\tt methods} and multiple {\it datasets}.  Initial - indicates the global result w/o re-ranking, while CG indicates results after applying CarGait. CarGait was applied on top of various gait recognition models, with wide range of performance levels and with publicly available code and checkpoints. Best results are in bold. (*) denotes unavailable code or checkpoint. (-) indicates no result for the corresponding setting.}
\label{tab:results}
\end{table*}

\begin{table*}[ht]
\centering
\small
\begin{tabular}{@{}c|c|cccc|cccc|cccc}
\toprule
  & & \multicolumn{4}{c|}{R1} & \multicolumn{4}{c|}{R5} & \multicolumn{4}{c}{mAP} \\
Method & Publication & KR & LBR & GCR & CG & KR & LBR & GCR & CG & KR & LBR & GCR & CG \\ \midrule
\tt{GaitPart}~\cite{fan2020gaitpart} & CVPR 20 & 26.5 & 23.3 & 26.0 & \textbf{29.5} & 42.7 & 47.1 & 45.7 & \textbf{48.5} & 21.50 & 18.24 & 21.68 & \textbf{22.73} \\
\tt{GaitGL}~\cite{lin2021gait} & ICCV 21 & 26.0 & 27.3 & 22.4 & \textbf{31.8} & 42.4 & 48.0 & 41.6 & \textbf{51.0} & 20.83 & 18.69 & 18.10 & \textbf{23.55} \\
\tt{GaitSet}~\cite{chao2019gaitset} {} & AAAI 19 & 34.8 & 33.0 & 35.7 & \textbf{41.5} & 53.1 & 60.4 & 56.4 & \textbf{62.1} & 30.26 & 26.92 & 30.53 & \textbf{32.97} \\
\tt{GaitBase}~\cite{fan2023opengait} & CVPR 23 & 60.0 & 63.8 & 63.1 & \textbf{66.1} & 77.6 & 82.7 & 79.2 & \textbf{82.8}  & \textbf{57.78} & 51.43 & 53.12 & 57.66 \\
\tt{DGV2-P3D}~\cite{fan2023exploring} {} & ArXiv 23 & 65.8 & 58.7 & 74.2 & \textbf{75.1} & 83.5 & 86.6 & 87.0 & \textbf{87.5} & 65.71 & 54.48 & 65.28 & \textbf{66.89} \\
\tt{SwinGait3D}~\cite{fan2023exploring} {}& ArXiv 23 & 66.7 & 64.0 & 74.1 & \textbf{76.3} & 83.6 & 88.1 & 86.3 & \textbf{88.6} & 66.79 & 57.79 & 64.03 & \textbf{67.59} \\
\tt{SG++}~\cite{fan2024skeletongait} {}& AAAI 24 & 69.7 & 61.7 & 76.1 & \textbf{78.1} & 85.6 & 90.2 & 89.6 & \textbf{90.4} & 70.30 & 58.99 & 68.72 & \textbf{70.86} \\
\bottomrule
\end{tabular}

\caption{Rank-$K$ accuracy [\%] and mAP on {\it Gait3D} dataset~\cite{Zheng_2022_CVPR} for different re-ranking methods: k-reciprocal (KR)~\cite{zhong2017re}, LBR~\cite{luo2019spectral}, and GCR~\cite{zhang2023graph}, compared to CarGait (CG). Best results are in bold.}
\label{tab:reranking_comparison_g3d}
\end{table*}
\begin{table}[ht]
\centering
\small
\begin{tabular}{@{}c|cccc}
\toprule
Method & Initial & KR & LBR & CG \\ \midrule
\tt{GaitPart}~\cite{fan2020gaitpart} & 88.5 & 68.4 & 80.6 & \textbf{89.1} \\
\tt{GaitGL}~\cite{lin2021gait} & 89.7 & 71.9 & 89.2 & \textbf{89.8} \\
\tt{GaitSet}~\cite{chao2019gaitset} {} & 87.1 & 65.5 & 70.3 & \textbf{87.5} \\
\tt{GaitBase}~\cite{fan2023opengait} & 90.8 & 72.9 & 74.4 & \textbf{91.1} \\
\tt{DGV2-P3D}~\cite{fan2023exploring} {} & 91.9 & 76.4 & 77.3 & \textbf{92.0} \\
\bottomrule
\end{tabular}

\caption{Rank-$1$ accuracy [\%] on {\it OU-MVLP} dataset~\cite{takemura2018multi} for different re-rankers: k-reciprocal (KR)~\cite{zhong2017re} and LBR~\cite{luo2019spectral}, compared to CarGait (CG). Initial - indicates the global result w/o re-ranking. This dataset is distinguished by having only a single positive sample in the gallery, making re-ranking particularly challenging. Best results are in bold. GCR~\cite{zhang2023graph} is excluded due to excessive runtime.}
\label{tab:reranking_comparison_oumvlp}
\end{table}
\begin{table}[ht]
\centering
\small
\begin{tabular}{@{}c|c|c}
\toprule
Method & Publication & Inference time [msec] \\ \midrule
KR~\cite{zhong2017re} & CVPR 17 & 214 \\
LBR~\cite{luo2019spectral} & ICCV 19 & 19.81 \\
GCR~\cite{zhang2023graph} {} & TMM 23 & 1866 \\
\midrule
\textbf{CarGait} & - & \textbf{6.52} \\
\bottomrule
\end{tabular}

\caption{Inference time per probe for various re-ranking methods, evaluated on the top-10 list. Best result is in bold.}
\label{tab:inference_time_analysis}
\end{table}

\subsection{Ablation Study and Analysis}
\label{sec:ablation_study}
To demonstrate the efficacy of the proposed method and assess the impact of various key components, we conducted an extensive ablation study using {\tt SwinGait-3D} model trained on {\it Gait3D} dataset. The results are summarized in~\cref{tab:ablations}, with further analysis provided in the appendix. While our primary training objective is to optimize Rank-1 (R1), we also examine the impact on Rank-5 (R5).
\begin{table}[ht]
\centering
\small
\begin{tabular}{@{}l|llll}
\toprule
{\#} & Method & R1 & R5 & mAP \\
\midrule
\multicolumn{5}{c}{Architectural and loss components} \\ 
\midrule
\rowcolor{lightgray}
1 & Pre-trained model w/o re-ranker & 75.0 & 86.7 & 66.69 \\
2 & Pre-trained model w/ extra training & 75.0 & 86.9 & 66.87 \\
3 & Binary classification (Baseline)  & 
71.6 & 87.8 & 65.47 \\
\midrule
% 3 & No relative probe \rnote{Should be removed} & 70.5 & 84.7 \\
4 & w/o classification loss ($\alpha=0$) & 76.0 & 89.1 & 67.73 \\
%5 & w/o ranking loss & 76.0 & 87.7 \\
5 & w/o loss damping ($\beta=1$) & 75.5 & 88.2 & 67.27 \\
\midrule
\multicolumn{4}{c}{Hyperparameters} \\
\midrule
6 & Inference re-ranking factor $K=5$ & 76.5 & 86.7 & 67.30 \\
7 & Inference re-ranking factor $K=20$ & 
76.3 & 88.6 & 67.68 \\
8 & Dataset creation with $v=20$ & 76.3 & 88.3 & 67.45 \\
9 & Dataset creation with $v=40$ & 76.0 & 89.0 & 67.66 \\
\midrule
10 & \textbf{CarGait} & 76.3 & 88.6 & 67.59 \\
\bottomrule
\end{tabular}

\caption{Ablation study with {\tt SwinGait-3D} model trained on {\it Gait3D} dataset. Rank-1 (R1), Rank-5 (R5), and mAP are reported.}
\label{tab:ablations}
\end{table}

\begin{figure}[!t]
\centering
\includegraphics[width=0.9\linewidth]{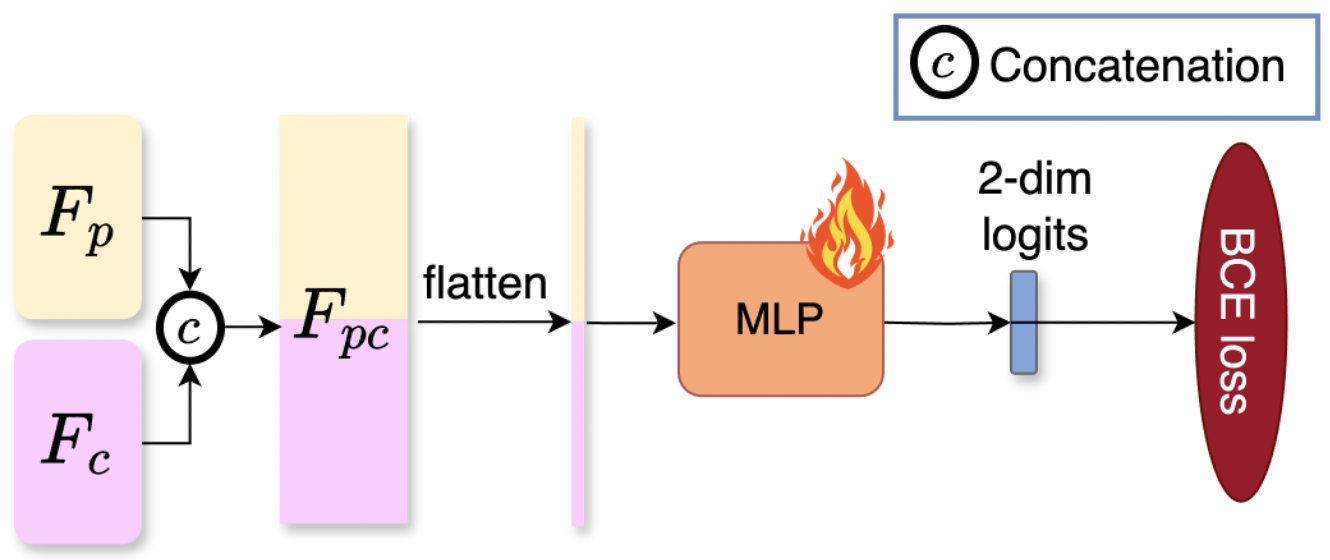}
\caption{An overview of the baseline binary classifier architecture presented in row-3 in~\cref{tab:ablations}. In this setup, there is no cross-attention module. Instead, the features $F_p$, $F_c$ are concatenated into $F_{pc}$ which is flattened and fed into an MLP to predict if it represents a positive or a negative pair (same or different identities).}
\label{fig:naive_arch_no_attention}
\end{figure}

\textbf{Baseline.} To show that additional training alone is insufficient to enhance performance, we continue training the global model for the same number of iterations used for CarGait. The result in~\cref{tab:ablations} (2nd row) suggests that the initial model (1st row) has already reached its full potential. 

To further highlight the advantages of our strip-wise cross-attention module, we also evaluate a naive baseline. In this baseline, the re-ranker is treated as a classifier trained on the top-$v$ global feature predictions, as illustrated in~\cref{fig:train_set_generation}.
To this end, we trained a binary classifier to identify whether a pair of probe and candidate sequences share the same identity (positives) or not (negatives). Practically, we trained a re-ranker without the cross-attention module, as illustrated in~\cref{fig:naive_arch_no_attention}. In this setup, we concatenated the two gait global features $F_p$, $F_c$ to $F_{pc}$, feeding it into an MLP that is further trained with BCE loss. At inference, the top $K$ pre-trained model predictions of a probe sequence are sorted in descending order based on their positive class scores. The result in~\cref{tab:ablations} (3rd row) shows a drop in Rank-1 (R1) performance compared to the initial reference state (1st row) for this baseline experiment. This decline can be attributed to naive interaction modeling and late fusion between the probe and candidates, whereas CarGait effectively learns internal part (strip) relationships, leading to improved recognition.

\textbf{Loss Components.} We conduct an ablation study on the classification loss to evaluate its impact during training. For this purpose, we train our model also without $\mathcal{L}_{CE}$ ($\alpha = 0$ in~\cref{eq:model_losses}). The result in row-4 confirms that incorporating the classification loss leads to additional improvements. Note that our optimization is focused on R1 and some components might negatively impact R5 or mAP. Then, in row-5, we present an experiment to evaluate the impact of our damping parameter $\beta$ in the loss function (see~\cref{eq:ranking_loss}). 

\textbf{Hyperparameters.} In rows 6-7, we evaluate the re-ranking performance in relation to $K$, the length of the top-ranked list on which re-ranking is performed during inference. The result in row 6 shows a slightly better R1, but CarGait with \( K=10 \) significantly enhances Rank-5 accuracy. While the mAP for \( K=20 \) (row 7) is slightly better, we chose \( K=10 \) as a fixed value across all experiments as a compromise between performance and runtime (see runtime analysis in the appendix). Finally, in rows 8-9, we examine the impact of the candidate set size during training (top-$v$). To this end, we create the training set with different values of $v$ ($20$ and $40$, instead of $30$). Although different values of $v$ could be optimized for each model and dataset, we fix $v=30$ across all experiments to ensure better generalization of our method. In the supplementary materials, we present a further analysis on $K$ and $v$ values.
\section{Conclusion}
\label{sec:conclusion}

In this paper, we introduce CarGait, Cross-Attention based Re-ranker for gait recognition. CarGait operates on pairs of probe and top-$K$ candidate feature maps, obtained from a pre-trained single-stage model, and introduces a new approach to generate {\it conditioned} representations for each probe-candidate pair. It then calculates the distances based on these updated representations to re-rank the original list, thereby improving Rank-1 and also Rank-5 accuracy. To this end, CarGait employs a cross-attention module that captures detailed correlations between pairwise probe-candidate body strips through their corresponding spatio-temporal feature maps. We evaluate CarGait with multiple models and across various datasets, demonstrating consistent performance improvements after re-ranking, and superior results over existing re-rankers. We believe this study marks an advancement for modern gait recognition, by exploring re-rankers, a field that has received very limited attention in the past.

{
    \small
    \bibliographystyle{ieeenat_fullname}
    \bibliography{main}
}
\clearpage
\maketitlesupplementary

\section{Hyperparameters}
\label{sec:sup_hyperparameters}

In the paper, we present an ablation study on the hyperparameters used in CarGait, $K$ and $v$. Both are fixed across all the experiments shown in the paper. The inference re-ranking factor $K$, which is the length of top-ranked list on which re-ranking is performed during inference, is set to $10$. The size of the candidate set in training, $v$, is set to $30$. Here, we present a more detailed analysis on $K$ and $v$, using the {\tt SwinGait-3D} model~\cite{fan2023exploring} trained on {\it Gait3D} dataset~\cite{Zheng_2022_CVPR}.

\begin{figure}[t]
\centering
\includegraphics[width=0.8\linewidth]{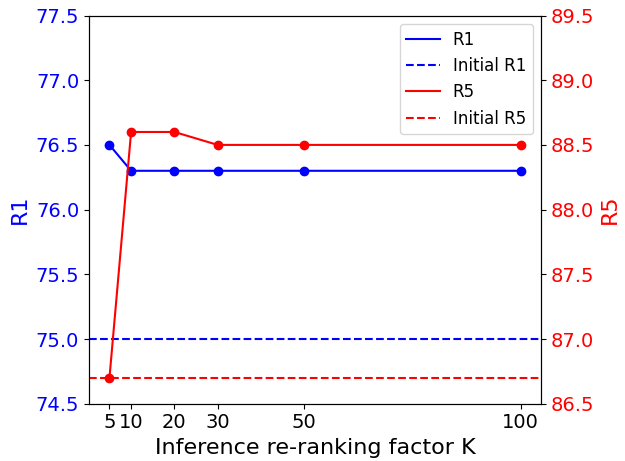}
\caption{Ablation study on CarGait inference re-ranking factor $K$, with {\tt SwinGait-3D} model trained on {\it Gait3D} dataset. Rank-1 (R1) and Rank-5 (R5) results are shown in blue (left-hand side) and red (right-hand side), respectively. The dashed lines indicate the initial single-stage model performance, while the solid lines represent the results after CarGait re-ranking.}
\label{fig:ablation_study_K}
\end{figure}

\begin{figure}[t]
\centering
\includegraphics[width=0.8\linewidth]{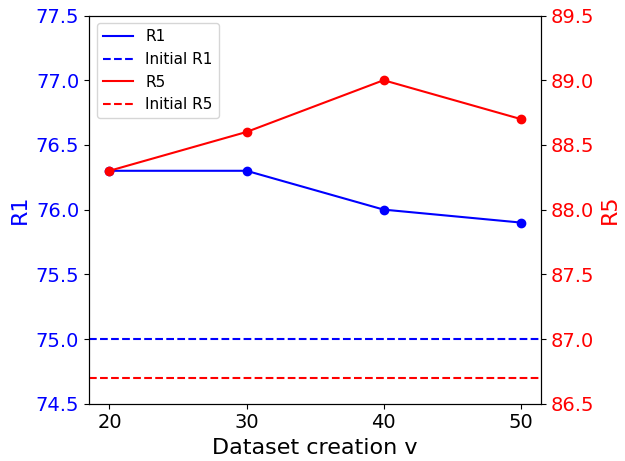}
\caption{Ablation study on CarGait training dataset creation $v$, with {\tt SwinGait-3D} model trained on {\it Gait3D} dataset. Rank-1 (R1) and Rank-5 (R5) results are shown in blue (left-hand side) and red (right-hand side), respectively. The dashed lines indicate the initial single-stage model performance, while the solid lines represent the results after CarGait re-ranking.}
\label{fig:ablation_study_v}
\end{figure}

As shown in~\cref{fig:ablation_study_K}, CarGait improvements are consistent across different values of $K$, with minor changes in R1 (higher for $K=5$), and in R5 (higher for $K=10$ and $K=20$). ~\Cref{fig:ablation_study_v} presents an ablation study on the top-$v$ candidates (per probe) used to construct the training set, as illustrated in the paper (Method section). Compared to the inference factor $K$, the performance variations here are more pronounced. Nevertheless, for all values of $v$ shown (solid lines in~\cref{fig:ablation_study_v}), both Rank-1 and 5 accuracy consistently surpass the initial state (dashed lines). 
As mentioned, CarGait with $v=30$ has been selected as a fixed hyperparameter for all {\tt models} and {\it datasets} to ensure better generalization of our method. That is, even though, in some cases, it might not be the optimal choice.

\section{Runtime and Memory Analysis}
\label{sec:sup_runtime_analysis}

\begin{figure}[!t]
\centering
\includegraphics[width=\linewidth]{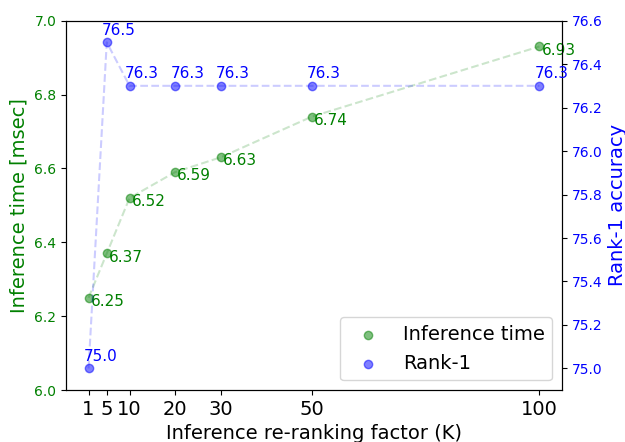}
\caption{CarGait runtime analysis per probe with varying values of $K$. The results were obtained on a single A100 GPU using the {\tt SwinGait-3D} model and the {\it Gait3D} dataset. The inference time per probe [in msec] is shown in green (left-hand side), while Rank-1 (R1) performance is depicted in blue (right-hand side).}
\label{fig:runtime_analysis}
\end{figure}

We provide a detailed inference runtime analysis of CarGait re-ranker across different $K$ values in~\cref{fig:runtime_analysis}. Our method involves a certain level of complexity. However, in practice, the inference overhead is only $\sim$6.5 [msec] on a single A100 GPU with $K=10$\footnote{For comparison, the first-stage global retrieval takes 0.1 [msec] per probe.}. As mentioned in the paper, the re-ranker size is influenced by the feature map dimensions obtained by the single-stage model. Generally, the number of trainable parameters varies from 2.07M to 9.21M. At inference, the re-ranker has 0.4M parameters at most (compared to the size of single-stage gait models that varies from 2.5M to 13M). Training the model on four A100 GPUs with 40 [GB] of RAM takes approximately 16 hours.

\begin{table*}[ht]
\centering
\small
\begin{tabular}{@{}c|c|cccc|cccc}
\toprule
  & & \multicolumn{4}{c|}{R1} & \multicolumn{4}{c}{R5} \\
Method & Publication & KR & LBR & GCR & CG & KR & LBR & GCR & CG \\ \midrule
\tt{GaitPart}~\cite{fan2020gaitpart} & CVPR 20 & 44.2 & 41.2 & 48.6 & \textbf{52.5} & 60.8 & 65.5 & 65.3 & \textbf{67.5} \\
\tt{GaitSet}~\cite{chao2019gaitset} {} & AAAI 19 & 44.7 & 41.2 & 48.6 & \textbf{52.0} & 62.6 & 65.5 & 65.3 & \textbf{68.0}  \\
\tt{GaitBase}~\cite{fan2023opengait} & CVPR 23 & 57.1 & 51.5 & 60.4 & \textbf{67.2} & 72.2 & 74.9 & 74.8 & \textbf{78.5} \\
\tt{DGV2-P3D}~\cite{fan2023exploring} {} & ArXiv 23 & 74.6 & 64.3 & 77.4 & \textbf{79.2} & 86.2 & 86.5 & 87.6 & \textbf{88.7} \\
\tt{SG++}~\cite{fan2024skeletongait} {}& AAAI 24 & 84.2 & 80.4 & 86.0 & \textbf{88.2} & 93.0 & 93.4 & 93.0 & \textbf{94.6} \\
\bottomrule
\end{tabular}

\caption{Rank-$K$ accuracy [\%] on {\it GREW} dataset~\cite{Zhu_2021_ICCV} for different re-ranking methods: k-reciprocal (KR)~\cite{zhong2017re}, LBR~\cite{luo2019spectral}, and GCR~\cite{zhang2023graph}, compared to CarGait (CG). Best results are in bold.}
\label{tab:reranking_comparison_grew}
\end{table*}

\section{Experiments on GREW}
\label{sec:rerankers_comparison_grew}

In the paper (Evaluation section), we demonstrate CarGait's superiority over the existing re-ranking methods~\cite{luo2019spectral, zhong2017re, zhang2023graph} on the {\it Gait3D}~\cite{Zheng_2022_CVPR} and {\it OU-MVLP}~\cite{takemura2018multi} datasets. Here, we provide an additional comparison on the {\it GREW}~\cite{Zhu_2021_ICCV} dataset. As shown in~\cref{tab:reranking_comparison_grew}, CarGait surpasses existing re-rankers across all five {\tt methods} in both Rank-1 and Rank-5 accuracy.

\section{Additional Experiments}
\label{sec:additional_experiments}

In Table 1 of the paper, we exclude results for settings where checkpoints are unavailable. To enrich our evaluation, we independently trained the {\tt SwinGait-3D} model on two datasets, using the official OpenGait implementation~\cite{Fan_2023_CVPR}. Although in {\it GREW} we were not able to reproduce the exact paper results, we provide CarGait performance gains on both datasets (see~\cref{tab:additional_results}).
\begin{table}[ht]
\centering
\footnotesize
\resizebox{\linewidth}{!}{%
\begin{tabular}{@{}c|cccc|cc@{}}
\toprule
 & \multicolumn{4}{c|}{\it GREW} & \multicolumn{2}{c}{\it OU-MVLP} \\
 & \multicolumn{2}{c}{R1} & \multicolumn{2}{c|}{R5} & \multicolumn{2}{c}{R1} \\
Method & Initial & CG & Initial & CG & Initial & CG \\ \midrule
\tt{SwinGait3D}~\cite{fan2023exploring} & 78.7 & \textbf{79.5} & 88.6 & \textbf{89.2} & 91.1 & \textbf{91.3} \\
\bottomrule
\end{tabular}%
}
\caption{Rank-$K$ accuracy [\%] on additional settings.}
\label{tab:additional_results}
\end{table}

\section{Verification}
\label{sec:verification}
In our evaluation, we adopt Rank-$K$ and mAP, which are the standard metrics in gait recognition. Here, we highlight the relevance of an important complementary metric: \textbf{verification} performance, measured by TPR@FPR. This metric evaluates how reliably and securely a system can verify identities under strict error restrictions, which may be a critical requirement in real-world scenarios.

To this end, we use the {\it Gait3D} dataset (which includes multiple positives per probe), and the top $K=1000$ single-stage predictions, to calculate TPR@FPR=1e-2 - before and after re-ranking. The results in~\cref{tab:verification} demonstrate consistent gains achieved by CarGait.

\begin{table}[ht]
\centering
\footnotesize
\resizebox{0.65\linewidth}{!}{%
\begin{tabular}{@{}c|cc}
\toprule
Method & Initial & CG \\
\midrule
\tt{GaitPart}~\cite{fan2020gaitpart}     & 21.0 & \textbf{21.9} \\
\tt{GaitGL}~\cite{lin2021gait}       & 21.9 & \textbf{27.5} \\
\tt{GaitSet}~\cite{chao2019gaitset}      & 27.1 & \textbf{39.3} \\
\tt{GaitBase}~\cite{fan2023opengait}     & 52.9 & \textbf{59.5} \\
\tt{DGV2-P3D}~\cite{fan2023exploring}   & 65.3 & \textbf{67.4} \\
\tt{SwinGait3D}~\cite{fan2023exploring} & 67.9 & \textbf{69.3} \\
\tt{SG++}~\cite{fan2024skeletongait}       & 69.7 & \textbf{73.7} \\
\bottomrule
\end{tabular}%
}
\caption{CarGait enhancements in verification performance (TPR@FPR=1e-2) on the {\it Gait3D} dataset~\cite{Zheng_2022_CVPR} with $K=1000$.}
\label{tab:verification}
\end{table}

\section{Ablations}
\label{sec:ablations}
Table 5 in the paper presents an ablation study using the {\tt SwinGait-3D} model trained on the {\it Gait3D} dataset. Here, we extend this analysis with additional ablations focused on the cross-attention module. CarGait employs a single cross-attention block with 8 heads. As shown in~\cref{tab:attention_ablation}, modifying the number of attention blocks or heads results in only a minor effect on performance.

\begin{table}[ht]
\centering
\footnotesize
\setlength{\tabcolsep}{7pt}
\renewcommand{\arraystretch}{1.05}
\begin{tabular}{@{}cccccc@{}}
\toprule
Method & \#Heads & \#Blocks & R1 & R5 & mAP \\
\midrule
\rowcolor{lightgray}
Initial     & -- & -- & 75.0 & 86.7 & 66.69 \\
H = 4   & 4  & 1  & 76.6 & 88.8 & 67.61 \\
H = 16  & 16 & 1  & 76.2 & 89.2 & 67.84 \\
B = 2   & 8  & 2  & 76.1 & 89.2 & 67.77 \\
B = 4   & 8  & 4  & 76.3 & 88.8 & 67.65 \\
\textbf{CarGait} & 8  & 1  & 76.3 & 88.6 & 67.59 \\
\bottomrule
\end{tabular}
\caption{Cross-attention ablations with {\tt SwinGait-3D} model and {\it Gait3D} dataset. Rank-1 (R1), Rank-5 (R5), and mAP are reported.}
\label{tab:attention_ablation}
\end{table}

\end{document}